\documentclass[9.5pt,journal,finalsubmission,compsoc]{IEEEtran}
\usepackage[left=0.5in,top=0.5in,bottom=0.5in,right=0.5in]{geometry}

\usepackage[pdftex]{graphicx}
\usepackage[cmex10]{amsmath}
\usepackage{amssymb}
\usepackage{amssymb}
\usepackage{theorem}
\usepackage{verbatim}
\usepackage{url}
\usepackage{wrapfig}
\usepackage{algorithm}
\usepackage{algorithmic}
\usepackage{boxedminipage}
\usepackage{float}
\usepackage{cite}
\usepackage{placeins} 

\usepackage[format=plain, labelsep=period,font={footnotesize,sf}]{caption}
\usepackage{subfig} 
\numberwithin{equation}{section}
\newcommand{\pd}[2]{ \frac{\partial #1}{\partial #2} }

\DeclareMathOperator*{\argmin}{argmin}
\newcommand{\eqnref}[1]{(\ref{#1})}

\newcommand{\figref}[1]{Fig. \ref{#1}}
\newcommand{\secref}[1]{Section \ref{#1}}
\newcommand{\Div}[1]{\nabla \cdot #1}
\newcommand{\grad}[1]{\nabla #1}

\newcommand{\norm}[1]{||#1||}
\newcommand{\transp}{^\mathsf{T}}  

\begin{document}
%
\title{Improved Edge Awareness in\\Discontinuity Preserving Smoothing}
%
%
%
%

\author{Stuart B. Heinrich and Wesley E. Snyder
\IEEEcompsocitemizethanks{\IEEEcompsocthanksitem S. Heinrich is
with the Department of Computer Science, North Carolina State
University, Raleigh,
NC, 27606.\protect\\
E-mail: sbheinri@ncsu.edu
}
}

\IEEEcompsoctitleabstractindextext{%
\begin{abstract}
Discontinuity preserving smoothing is a fundamentally important
procedure that is useful in a wide variety of image processing
contexts.  It is directly useful for noise reduction, and
frequently used as an intermediate step in higher level
algorithms.  For example, it can be particularly useful in edge
detection and segmentation.  Three well known
algorithms for discontinuity preserving smoothing are nonlinear
anisotropic diffusion, bilateral filtering, and mean shift
filtering.  Although slight differences make them each better
suited to different tasks, all are designed to preserve
discontinuities while smoothing.  However, none of them satisfy
this goal perfectly: they each have exception cases in which
smoothing may occur across hard edges.  The principal contribution of this paper is the identification of a property we call edge awareness that should be satisfied by any discontinuity preserving smoothing algorithm.  This constraint can be incorporated into existing algorithms to improve quality, and usually has negligible changes in runtime performance and/or complexity.  We present modifications necessary to augment diffusion and mean shift, as well as a new formulation of the bilateral filter that unifies the spatial and range spaces to achieve edge awareness.
\end{abstract}

}

\maketitle

\IEEEdisplaynotcompsoctitleabstractindextext

%
\IEEEpeerreviewmaketitle

\section{Introduction}
\IEEEPARstart{D}{iscontinuity} preserving algorithms attempt to
smooth an image while preserving crisp edges.  They are useful for
many tasks, such as noise reduction, edge detection, and
segmentation.  Some very robust results have been obtained by
treating the smoothing process as a global optimization problem,
where the objective is to recover a smooth image that has high
fidelity to the original corrupted image.  Typically, the energy
function to be minimized is a summation over all pixels in the
image with two terms, one of which is minimized by a perfectly
smooth image, and the other by perfect fidelity to the original
corrupted image.

It has been shown that the solution that minimizes such a function
is also the maximal \emph{a posteriori} solution to the problem,
as defined by the prior and data terms \cite{HAN98}
\cite{SNYDER04} \cite{FELZENSZWALB06}.  Although this is an
NP-hard \cite{SCHARSTEIN02} non-convex optimization problem
\cite{BLACK94}, many novel algorithms have been proposed which can
efficiently find good global minima in polynomial time, including
Mean Field Annealing (MFA) \cite{HAN98} \cite{SNYDER04}
\cite{ZERUBIA93} \cite{WASSERSTROM73}, Graduated Non-Convexity
(GNC) \cite{BLAKE87}, Belief Propagation (BP)
\cite{FELZENSZWALB06}, and Graph Cuts (GC) \cite{BOYKOV01}
\cite{BOYKOV04} \cite{GREIG89}.  Earlier works have also identified specific line and outlier processes, further discussed in \cite{BLACK94}.

However, the global approaches have some significant
disadvantages: they are often difficult to generalize to
vector-valued images, they are generally very inefficient, they
often have trouble maintaining perfectly smooth gradients, and
fine structure may be lost.

In this paper, a few algorithms are discussed which have emerged
as some popular efficient alternatives to global optimization.  The algorithms we consider are variable conductance diffusion \cite{GROSSBERG84}
\cite{NORDSTROM} \cite{PERONA90} \cite{SNYDER04} \cite{BARASH00},
bilateral filtering \cite{TOMASI98} \cite{BARASH00}, and mean
shift filtering \cite{COMANICIU02}.

\begin{figure*}
\centering
\begin{tabular}{c}
\subfloat[Original]{\includegraphics[width=2in]{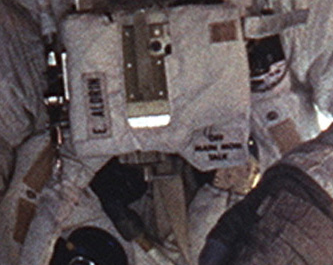}
}
\subfloat[Diffusion]{\includegraphics[width=2in]{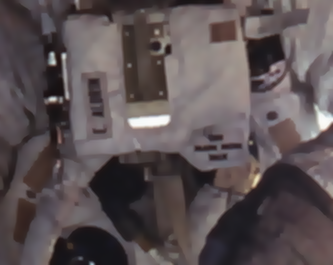} }\\
\subfloat[Bilateral]{\includegraphics[width=2in]{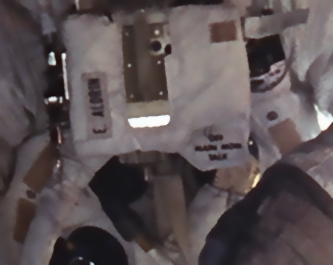}
} \subfloat[Mean
shift]{\includegraphics[width=2in]{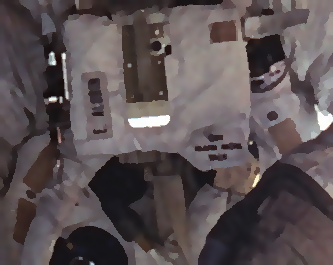}
}
\end{tabular}
\caption{Original (a) and filtered result of \emph{Apollo} image
using: (b) 5 iterations of diffusion with $\lambda=0.03$; (c)
regular bilateral filter with $\sigma_S=3$, $\sigma_R=45$; (d)
regular mean shift filter with $h_s=5$ and $h_r=45$.}
\label{apollo_test_old}
\end{figure*}

An example of filtering using each of these methods on the noisy
\emph{Apollo} image is shown in \figref{apollo_test_old}.  As can
be seen from this test image, each method can be effectively used
to reduce noise while preserving sharp edges.  Although these
algorithms perform the same fundamental task, there is no clear
winner between them -- they all perform well, and individual
differences between them (discussed in more detail in the
following sections) are significant enough that each has found a
niche where it may be superior to the others.

The mean shift filter produces more crisp lines in homogeneous
regions because it was originally designed for a somewhat ulterior
purpose -- it is an adaptation of mean shift clustering for
fine-scale segmentation of color images.  Thus, smooth gradients
are avoided, because pixel colors represent segments, and it is
only a coincidence that this also appears to be an effective
mechanism for discontinuity preserving smoothing.

The bilateral filter and diffusion methods appear to produce
results of similar quality, but they are not equal; the bilateral
filter can produce a greater contrast between sharp edges and very
smooth regions, and is more efficient on massively parallel
architectures.  Conversely, diffusion is more efficient on single
processor machines, and has a more physically meaningful result
because it closely approximates the physical diffusion of
densities through a medium.

\begin{figure}[H]
\centering \subfloat[Original]{
\includegraphics[width=0.8in]{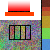} }
\subfloat[Diffusion]{\includegraphics[width=0.8in]{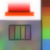}
}
\subfloat[Bilateral]{\includegraphics[width=0.8in]{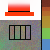}}
\hspace{0.3pt} \subfloat[Mean
shift]{\includegraphics[width=0.8in]{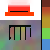}
} \caption{Synthetic \emph{Challenge} (a) test image exhibiting
some properties that cause existing methods to perform poorly: (b)
variable conductance diffusion blurs hard edges (intentionally exaggerated here for illustration purposes); (c) the bilateral
filter loses saturation and fails to smooth in high-noise regions;
(d) mean shift mangles some of the edges.} \label{difficult}
\end{figure}

We make no attempt to compare or rank the aforementioned
algorithms against each other.  However, we have noticed that that
there are certain limitations to each algorithm which can result
in unwanted effects, and examples of these issues are demonstrated
on a specially designed \emph{Challenge} test image, shown in
\figref{difficult}.

There are several notable features of this test image.  In the
upper region, there is a gradient going from the white background
to a solid red color, adjacent to an orange rectangle, which is
separated by a thin black line.  Because the red and orange
regions are clearly separated, we claim that smoothing should not
occur across these regions.

In the bottom half of the image there are four black rectangles
each containing noise with a different dominant color: red, green,
yellow, or purple.  We claim that a smoothing algorithm should
reduce the noise within these rectangles leaving each one filled
with a solid color, and also preserve the crisp edges of the
rectangles.

Surrounding the rectangles is more brightly colored noise, but the
noise is not uniform -- the area is divided into blobs with
distinctly dominant colors.  We claim that a smoothing algorithm
should reduce this to a smoothly varying colored surface.

In the right region, there are 6 distinctly colored rectangles
containing low levels of zero-mean noise.  This presents a
particularly simple task for a discontinuity preserving smoothing
algorithm, and is shown here as somewhat of a control.

From \figref{difficult}b, we see that successive iterations of
diffusion erodes even the hardest edges, due to the Gaussian blurring of derivatives (we have used a blurring radius that makes this effect noticeable, although it could be tweaked to produce a nicer image).  From \figref{difficult}c, we
see that the bilateral filter reduces saturation, produces an unwanted
new color by blending the red and orange sections, and has
trouble smoothing the highly noisy colored areas.  From
\figref{difficult}d, we see that the mean shift filter mangles the
black border edge features.

In this paper, we investigate the cause of these unwanted effects,
and show that most of them can be related to the violation of a
certain principle which we refer to as \textbf{edge awareness}.
We propose modifications to correct each algorithm with little or
no changes in time complexity, and demonstrate that these
modifications can significantly improve visual quality of
smoothing.  Specifically, we will show that

\begin{enumerate}
    \item Edge awareness is a concept that can be used to reduce estimation bias in image restoration.
    \item Diffusion is naturally edge aware, but becomes progressively less so with each iteration.  This can be fixed to preserve crisp edges after successive iterations, at minor additional expense and no change in time complexity (as shown in \secref{diffusion_improve}).
    \item Edge awareness can be incorporated into the concept of the bilateral filter in a very intuitive way to produce a novel algorithm that unifies the spatial and range spaces.  Our modification multiples the time complexity as compared to a standard bilateral filter by $O(\log d)$, for a neighborhood of size $d$ (as shown in \secref{bilateral_improve}).
    \item Edge awareness can be partially incorporated into the mean shift filter with minimal change in running time by preventing mean shift paths from crossing strong intensity boundaries (as shown in \secref{ms_ea}).  This prevents unwanted artifacts from appearing.
\end{enumerate}

\section{Edge Awareness}

In order to be called discontinuity or edge preserving, a
smoothing algorithm needs only to smooth areas without edges more
than the areas with edges.  This leaves considerable flexibility
in interpretation, and many algorithms satisfying this criterion
exist.  However, we argue that a more restrictive criterion may be
better used to describe an unbiased discontinuity preserving
smoothing algorithm, which we refer to as edge awareness.

Specifically, the criterion for edge awareness is:

\begin{quote}
\emph{The influence that a pixel in the input image has on the
output at another pixel location should be inversely related to
the strength of the largest edge(s) separating those pixels in the
input image.}
\end{quote}
\vspace{0.1in}

This criterion has statistical rationale behind it for noise
removal.  The process of smoothing can be equated to estimating
the sample mean of the color reflectance distribution at each
pixel, and discontinuities in an image are reliable indicators of
a change in the underlying reflectance distribution.  Without any
\emph{a priori} knowledge of the distributions contained within an
image, disjoint regions must be assumed to come from disjoint
distributions, and samples from one distribution cannot possibly
help in estimating the mean of another distribution.  Therefore,
pixels which are separated by a discontinuity, and hence not
likely to come from the same distribution, should not influence
each other.

\begin{figure}[H]
\centering
\includegraphics[width=3.5in]{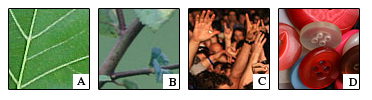}
\caption{Color distributions in real images: (a) the veins of a
leaf separate regions of green having the same underlying
distribution; (b) a foreground twig separates regions of a
background out-of-focus, having a uniform distribution; (c) a
crowd of people with many similar but differently colored body
parts; (d) a collection of buttons with similar but different
distributions of red. } \label{distribs}
\end{figure}

Arguably, there are many cases where a single reflectance
distribution could be responsible for generating samples at
disconnected regions in the image.  For example, in the regions
separated by veins in a leaf, or the out of focus background in a
photograph separated by occluding foreground objects (see images A
and B in \figref{distribs}).  In these cases the quality of the
sample mean could be improved by using the samples from both
distributions (eg, smoothing between disjoint regions).

However, an unbiased smoothing algorithm should not assume that
two distributions are the same just because their sample means are
similar, because this is frequently not the case.  As an example,
consider a crowd of people: people tend to have similar skin
colors which could appear close together in the image, but a
smoothing algorithm should not change a person's skin color by
blending them together (see examples C and D in
\figref{distribs}).

A smoothing algorithm that is not edge aware imposes the
additional, often unwanted, assumption that distributions having
similar means are the same distribution.  Doing so can result in
the creation of novel colors that did not exist in the original
image, which is counter productive if the purpose of smoothing is
to recover a better estimate of the underlying process by removing
noise (as it frequently is).

\section{Variable Conductance Diffusion}

Variable conductance diffusion \cite{GROSSBERG84} \cite{NORDSTROM}
\cite{PERONA90} \cite{SNYDER04} \cite{BARASH00} has long been used
as a method of discontinuity preserving smoothing for images.  It
is inspired by the diffusion equation \eqnref{diffusion}, and uses
a heuristic measure of conductivity $c$ that is somehow based on
the local image derivative magnitude.

\begin{equation} \label{diffusion}
\pd{f(x,y)}{t} = \Div{ (C(x,y) \grad{f(x,y)} )}
\end{equation}

For gray level images, the diffusion equation directly translates
into an iterative discontinuity preserving smoothing algorithm
using finite difference approximations to the derivatives, where
$f(x,y)$ is image intensity, and conductivity is a decreasing
function of local gradient magnitude \cite{BARASH00}
\cite{PERONA90},

\begin{equation}
C(x,y) = g( ||\grad{f(x,y)}|| )
\end{equation}

In our implementation, we deviate slightly by using a
vector-valued conductivity,

\begin{equation}
\begin{split}
\pd{f(x,y)}{t} &= \Div{ (C(x,y) \transp \grad{f(x,y)} )} \\
&= \pd{}{x} \left( c_x(x,y) \pd{f(x,y)}{x} \right) + \pd{}{y}
\left( c_y(x,y) \pd{f(x,y)}{y} \right)
\end{split}
\end{equation}

\noindent and we make smoothing in each dimension a function of
the partial derivative, so that $C(x,y) = [ c_x(x,y) c_y(x,y) ]
\transp$.  The components $c_x(x,y)$ (and correspondingly
$c_y(x,y)$) are defined by

\begin{equation}
c_x(x,y) = \exp\left( -\lambda \norm{ \pd{f(x,y)}{x} }^2 \right)
\end{equation}

For color images, the conductance in each channel can be made a
function of the overall intensity gradient.  Alternatively, Sapiro
\& Ringach \cite{SAPIRO96} have proposed a method based on the
local structure tensor.

VCD is sometimes confused with anisotropic diffusion, which is
almost identical but explicitly diffuses in the direction tangent
to the local edge \cite{SNYDER04}.  In practice, all of these
diffusion methods have similar performance.

The diffusion process attempts to smooth in a way that respects
intensity boundaries by making the amount of smoothing
related to the magnitude of the derivatives.  As a result,
colors should not be blurred together unless there exists some
path between them that does not significantly go against any local
image gradients.  Therefore, diffusion is edge aware in theory.

\subsection{Limitations of Diffusion}

Finite differences are first order approximations of the true derivative, and iterative use causes these approximation errors to become magnified.  One way to help mitigate this problem is to alternate between the causal and non-causal forms of the derivative on each iteration.  However, noisy images may contain a large degree of high frequency content so that this is generally not enough to eliminate artifacts.

In practice, a Gaussian blurring step is used before computing the
finite difference derivative approximations \cite{SAPIRO96}.  This naturally makes the derivatives more stable by considering a slightly lower frequency.  If noise is mostly high frequency, this also makes the derivatives more representative of the underlying signal.

Unfortunately, because edges are inherently high frequency, lower frequency derivatives will not be able to accurately represent crisp edges, which will lead to slight blurring of edges with each iteration (as in
\figref{difficult}b).  Although this blurring effect can be mitigated by using a very small sigma, it will always be present. Thus, the algorithm becomes progressively less aware of edges/discontinuities with each iteration, and will eventually break down and converge to smoothing everywhere.  In this paper, we intentionally use a blurring sigma that is large enough to be noticeable.

\section{The Bilateral Filter} \label{bilateral_filter}

The bilateral filter was developed by Tomasi \& Manduchi for
discontinuity preserving smoothing as a non-iterative alternative
to anisotropic diffusion \cite{TOMASI98}.  The primary claim for
improvement over anisotropic diffusion is that the iterations in
diffusion raise issues of stability, as small errors in the
derivative estimation can become magnified over time, producing a
less smooth image.  Additionally, iterative algorithms can be
inefficient for massively parallel architectures with limited
communication.

The bilateral filter overcomes these issues by considering a
single fixed neighborhood of influence around each pixel, and
estimating the influence of each neighbor by a combined spatial
and range weighting.  Specifically, each output pixel is replaced
by a doubly-weighted average using the spatial distance and range
(color) distance of the original neighbors.  For each pixel
coordinate $s$, the output color $J_s$ is given by

\begin{equation}
J_s = \frac{1}{k} \sum_{p \in \Omega} f(||p-s||) g(||I_p - I_s||)
I_p \label{bilateral}
\end{equation}

\noindent where $\Omega$ is the set of image pixels, $I_p$ is the
RGB intensity at pixel $p$ in the input image $I$, and $k$ is a
normalization term.  $f$ and $g$ could be any 1D kernel functions,
but in practice $g$ is a true Gaussian with standard deviation
$\sigma_R$ and $f$ is a truncated Gaussian (for computational
efficiency) with standard deviation $\sigma_S$.

Because it is not iterative and does not rely on derivatives, this
makes it more numerically stable, but the use of large overlapping
neighborhoods is also much more inefficient; unlike the Gaussian
blur, this filter cannot be separated into a horizontal and
vertical pass for increased efficiency.

\subsection{Limitations of the Bilateral Filter} \label{bilat_limit}

The bilateral filter is completely unaware of edges because neighboring pixels are averaged together
based only on their color similarity, with no regard to any edges
that might be separating them.  As the spatial radius gets larger,
the likelihood of two similarly colored pixels in the neighborhood
being separated by some kind of discontinuity grows very quickly.
This results in blurring across edges.  In addition to producing
novel colors, this can produce oddly surrealistic effects because
it often blurs the interior of foreground objects with the
background, as if it were a line drawing.

Even if the foreground and background colors are noticeably
different, a high noise content may necessitate using a large
color bandwidth, so that these colors are still blurred, resulting
in noticeable color bleeding across boundaries.  Also, opposing
bright colors with similar intensity may be blurred together,
resulting in muted grays and browns instead of the original
vibrant colors (this is demonstrated by the noisy background in
\figref{difficult}).  This effect will be referred to as
\emph{color cancellation}, and results in loss of saturation.

Some images are more prone to these issues than others.  If an
image does not contain nearby opposing colors, has a sparse
histogram, or has a very high signal to noise ratio, then the
bilateral filter may producing very visually appealing results.
However, this does not always mean that the results are accurate,
because color bleeding across boundaries will still result in
changing the average colors even if they still look good to a
human.

It should be noted that, in some contexts, blurring across
boundaries is considered an advantage rather than a limitation.
For example, if the objective is to generate a cartoon-like image
rather than produce a realistic smoothing.

\section{The Mean Shift Filter}

Mean shift clustering \cite{COMANICIU02} was originally developed
as a method of clustering multi-dimensional data using an
iterative hill-climbing technique to assign each pattern to the
nearest mode in a kernel density estimate of the global pattern
distribution.  A kernel density estimate is essentially an
estimate of the probability mass function that generated a set of
data points, and can be recovered by convolving a Gaussian with
each data point.

Conceptually, the mean shift vector is a step along the gradient
of the kernel density estimate towards the nearest mode.  A
multi-dimensional mean value is initialized by the data point
itself, and then a new multi-dimensional mean that is closer to
the mode is computed by the average of all patterns weighted by
their distance to the current multi-dimensional mean.

Formally, the mean shift vector $m(x)$ at point $x$, which is the
vector towards the next point on the path to the mode
\cite{COMANICIU02}, is usually written as,

\begin{equation}
m(x) = \frac{\sum_{i=1}^n x_i g\left( ||\frac{x-x_i}{h}||^2
\right)}{\sum_{i=1}^n g\left( ||\frac{x-x_i}{h}||^2 \right)} - x
\label{mean_shift_vec}
\end{equation}

\noindent for a set of $n$ patterns $x_i, ..., x_n$ and spatial
bandwidth $h$.  $g$ is any 1D kernel, but a Gaussian is typically
used.

Mean shift clustering is particularly appealing because it is not
only one of the few clustering algorithms that is not biased
towards hyper-spherical or even hyper-ellipsoidal clusters, but is
actually non-parametric (meaning that it is completely unbiased to
the distribution) and still quite robust.

Each pixel in an image can be thought of as a 5-dimensional point
characterized by $(x,y,r,g,b)$, and mean shift clustering can be
used to find the modes of that distribution.  When it is used in
this context, it is known as \emph{mean shift filtering}
\cite{COMANICIU02}, although it is more effective to compute
distance as the sum of Euclidean distances in the spatial and
range components separately, as opposed to computing the
5-dimensional Euclidean distance.  Therefore, two parameters are
needed to specify the bandwidth (or equivalently, standard
deviation) in those two domains.

Although mean shift clustering is effective for clustering many
generic data sets, in the context of image processing mean-shift
filtering tends to find too many modes for a suitable segmentation
-- but it does make a good step in the right direction.  One of
the most popular method of segmenting images, \emph{mean shift
segmentation}, first uses the mean shift filter, and then uses the
union-find algorithm with region adjacency graphs to cluster the
filtered image pixels (in the `fusion' step) \cite{COMANICIU02}.

\subsection{Limitations of Mean Shift}

Despite the very different form presented in
\eqnref{mean_shift_vec} from \eqnref{bilateral}, the mathematics
of mean shift filtering are actually quite similar to that of
bilateral filtering, except: where the bilateral filter used a
bilateral weighting of the local neighborhood as a
\emph{replacement} to the pixel at the origin, the mean shift
filter uses a bilateral weighting of the local neighborhood to
compute \emph{the next step} towards the local kernel density
estimate of the multivariate image intensity distribution.

In fact, if $s_i$ is used to denote the $i$th shifted position of
the mean starting at pixel $s$, then mean shift filtering can be
written in the same form as the bilateral filter, because the
bilateral filter is essentially identical to 1 iteration of the
mean shift filter:

\begin{equation}
s_{i+1} = \frac{1}{k} \sum_{p \in \Omega} f(||p-s_i^S ||) g(||I_p
- s_i^R||) I_p \label{ms_bilateral}
\end{equation}

\noindent where $s_i^S$ is the spatial component of the mean, and
$s_i^R$ is the range component.  This similarity seems to have
been previously overlooked in the literature.

Because the mean shift algorithm shifts the location of the mean
on each iteration until a local mode of the distribution is found,
the range of influence is not restricted to pixels in the initial
neighborhood of a pixel -- theoretically, any pixel in the input
image could influence the color of any pixel in the filtered
output image.

In practice, the spatial bandwidth used to compute averages in
mean-shift filtering can be much smaller than for bilateral
filtering because the window moves iteratively.  Because images
contain many modes, the number of mean shift iterations is usually
not too large, and as a result mean-shift filtering can actually
be significantly more efficient than bilateral filtering for
comparable ranges of smoothing influence.  In addition, because
mean-shift filtering does hill-climbing on the underlying image
distribution, the mean shift path has a tendency to not cross
large boundaries in the image, which allows it to perform blurring
over larger effective ranges than a bilateral filter without
significant color bleeding across boundaries.

By replacing each pixel with the color of its nearest associated
mode, the mean-shift filter tends to reduce gradients in the
original image.  However, it also can create gradients that did
not previously exist between two regions if their color difference
is small relative to $h_r$.  If the filter is being used as a
precursor to clustering (such as is the case in the popular mean
shift segmentation algorithm \cite{COMANICIU02}), then reducing
gradients can be an important property because pixels on a
gradient should logically be clustered together.

Because the mean-shift path essentially follows the gradient
towards the mode, it tends to not cross boundaries in the original
image, but there is no guarantee that it won't.  As the spatial
bandwidth becomes larger, it becomes increasingly likely that it
will, for two reasons.  The first reason is that because the mean
shift uses a bilateral weighting, it may give high weighting to
pixels that are separated by an intensity border.  However, this
is less likely to be a significant problem because the spatial
bandwidth is generally much smaller when using mean shift than the
bilateral filter.

The second, more prominent reason, is that there is nothing
preventing the \emph{spatial mean} of similarly colored nearby
pixels from shifting to a location which has a drastically
different pixel color.  This happens a lot around non-convex
colored image features, such as the black rectangles in
\figref{difficult}d.  The most obvious conceptual example is a
donut shaped feature: suppose the mean shift vector is attempting
to converge to the mode of this distribution.  If the spatial
bandwidth is large relative to the donut, then the curvature of
the donut will cause the spatial mean to fall within the center of
the donut.

Because mean shift uses a multi-variate mean, this won't
\emph{immediately} change the color of the range component of the
mean.  However, after the spatial component converges to the
center of the donut, the color component will slowly converge to
the color of the interior of the donut because of the high
weighting from the spatial component of the bilateral weighting.
The end result of this effect is that a pixel which started out on
the donut is assigned to the mode of a completely different
distribution: that of the background behind the donut, and this is
a violation of edge awareness.

These problems essentially limit the range of bandwidths that can
be used with good results in mean shift, restricting it to
relatively small scale (high frequency) kernel density estimates.

\section{Edge Aware Modifications for\\ Diffusion} \label{diffusion_improve}

Our edge aware modification to diffusion is extremely simple:
rather than using a Gaussian blur to condition the derivatives, we
propose to use a bilateral filter (discussed in more detail in
\secref{bilateral_filter}).  Although we have already illustrated
some exceptions in which the bilateral filter does not properly
account for edges in \secref{bilat_limit}, these limitations are
only present when using a large spatial support.  The blurring
process here is only used to fix pixel discretization errors, so a
\emph{small fixed size} bilateral filter using only the adjacent
neighborhood is sufficient.  In all examples in this paper, we use
a $5 \times 5$ neighborhood with a spatial standard deviation of
0.5.  At this small scale, the limitations of the bilateral filter
(discussed in \secref{bilat_limit}) are irrelevant.

\subsection{Performance Analysis of Edge Aware Diffusion}

The original diffusion equation is extremely efficient for
sequential machines because, in each iteration, color is
propagated locally based on derivatives.  This eliminates the need
to consider large neighborhoods.  Each iteration can be done in
low $O(n)$ time.  Thus, for a range of influence of $i$ pixels,
the time complexity of the original algorithm is $O(n i)$.

The Gaussian blur is a linearly separable operator which can be
computed in two passes using one-dimensional kernels for improved
performance, whereas the bilateral filter used in the proposed
edge aware modified version cannot be.  However, because the
neighborhood considered by the bilateral filter is a small fixed
size, this can be treated as a constant.  Therefore, switching to
a bilateral filter does not change the time complexity of the
algorithm, and only slightly increases the overall constant
factor.

\subsection{Experimental Results of Edge Aware Diffusion}

\begin{figure*}
\centering
\begin{tabular}{c}
\subfloat[Froud]{\includegraphics[width=2in]{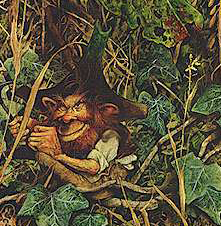}
}
\subfloat[Diffusion]{\includegraphics[width=2in]{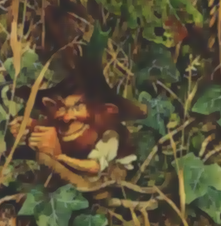}
}
\subfloat[Edge Aware Diffusion]{\includegraphics[width=2in]{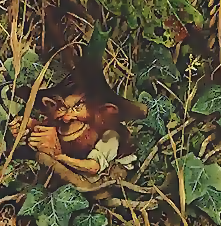} }\\
\subfloat[Paris]{\includegraphics[width=2in]{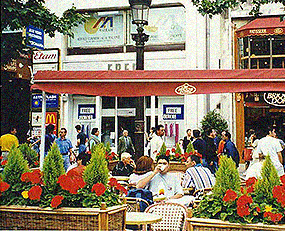}
}
\subfloat[Diffusion]{\includegraphics[width=2in]{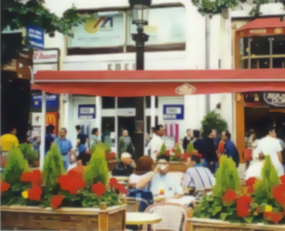}
}
\subfloat[Edge Aware Diffusion]{\includegraphics[width=2in]{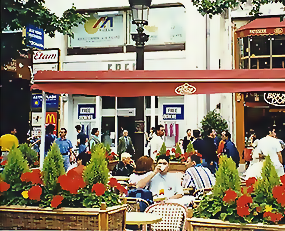} }\\
\subfloat[Nest]{\includegraphics[width=2in]{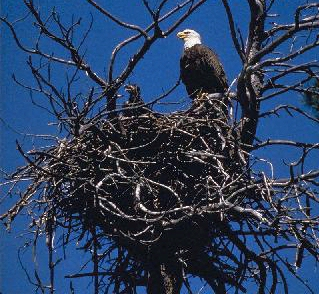}
}
\subfloat[Diffusion]{\includegraphics[width=2in]{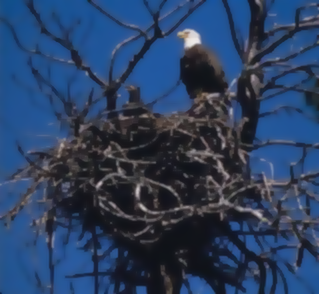}
}
\subfloat[Edge Aware Diffusion]{\includegraphics[width=2in]{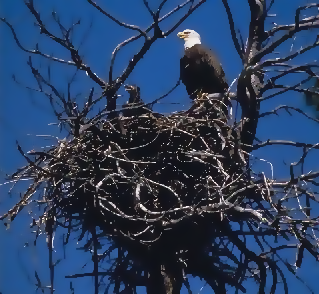} }\\
\subfloat[Egg]{\includegraphics[width=2in]{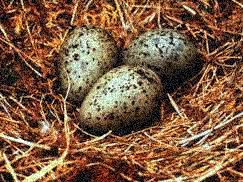}
}
\subfloat[Diffusion]{\includegraphics[width=2in]{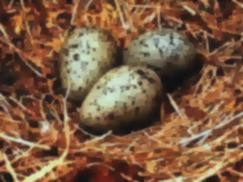}
}
\subfloat[Edge Aware Diffusion]{\includegraphics[width=2in]{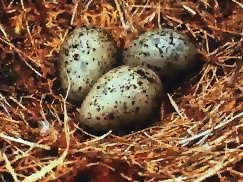} }\\
\end{tabular}
\caption{Comparison between nonlinear anisotropic diffusion
filtered results on input image (left column) using standard
algorithm (middle column) and edge aware modified algorithm (right
column).  $\lambda=0.002$ and iterations=5 for all tests.  For
smoothing prior to taking derivatives, $\sigma_S=0.5$, and in the
case of bilateral filter $\sigma_R=55$.} \label{vcd_test_compare}
\end{figure*}

The original and edge aware modified diffusion filters are tested
on a selection of images in \figref{vcd_test_compare}.  We have
intentionally used a blurring radius that is large enough to
exaggerate the smoothing out of edges in the original diffusion
algorithm.  This makes the improvements of our edge aware
diffusion algorithm, which preserves hard edges even with large
blurring radius, more noticeable.  It should be noted that a
better looking result could have been obtained using the regular
diffusion algorithm by using a smaller blurring radius.

\begin{figure*}
\centering
\begin{tabular}{c}
\subfloat[$\lambda=0.0005,
i=1$]{\includegraphics[width=1in]{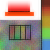}
} \subfloat[$\lambda=0.0005,
i=2$]{\includegraphics[width=1in]{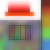}
} \subfloat[$\lambda=0.0005,
i=3$]{\includegraphics[width=1in]{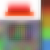}
} \subfloat[$\lambda=0.0005,
i=4$]{\includegraphics[width=1in]{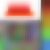}
}
\subfloat[$\lambda=0.0005, i=5$]{\includegraphics[width=1in]{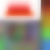} }\\
\subfloat[$\lambda=0.0105,
i=1$]{\includegraphics[width=1in]{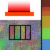}
} \subfloat[$\lambda=0.0105,
i=2$]{\includegraphics[width=1in]{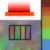}
} \subfloat[$\lambda=0.0105,
i=3$]{\includegraphics[width=1in]{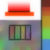}
} \subfloat[$\lambda=0.0105,
i=4$]{\includegraphics[width=1in]{vcd_lambda_0p010500_i4.png}
}
\subfloat[$\lambda=0.0105, i=5$]{\includegraphics[width=1in]{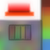} }\\
\subfloat[$\lambda=0.0500,
i=1$]{\includegraphics[width=1in]{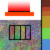}
} \subfloat[$\lambda=0.0500,
i=2$]{\includegraphics[width=1in]{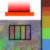}
} \subfloat[$\lambda=0.0500,
i=3$]{\includegraphics[width=1in]{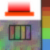}
} \subfloat[$\lambda=0.0500,
i=4$]{\includegraphics[width=1in]{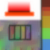}
} \subfloat[$\lambda=0.0500,
i=5$]{\includegraphics[width=1in]{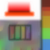}
}
\end{tabular}
\caption{Nonlinear anisotropic diffusion results on
\emph{Challenge} image using standard algorithm.}
\label{vcd_challenge}
\end{figure*}

\begin{figure*}
\centering
\begin{tabular}{c}
\subfloat[$\lambda=0.0005,
i=1$]{\includegraphics[width=1in]{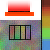}
} \subfloat[$\lambda=0.0005,
i=2$]{\includegraphics[width=1in]{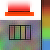}
} \subfloat[$\lambda=0.0005,
i=3$]{\includegraphics[width=1in]{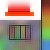}
} \subfloat[$\lambda=0.0005,
i=4$]{\includegraphics[width=1in]{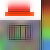}
}
\subfloat[$\lambda=0.0005, i=5$]{\includegraphics[width=1in]{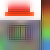} }\\
\subfloat[$\lambda=0.0105,
i=1$]{\includegraphics[width=1in]{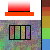}
} \subfloat[$\lambda=0.0105,
i=2$]{\includegraphics[width=1in]{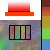}
} \subfloat[$\lambda=0.0105,
i=3$]{\includegraphics[width=1in]{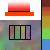}
} \subfloat[$\lambda=0.0105,
i=4$]{\includegraphics[width=1in]{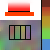}
}
\subfloat[$\lambda=0.0105, i=5$]{\includegraphics[width=1in]{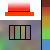} }\\
\subfloat[$\lambda=0.0500,
i=1$]{\includegraphics[width=1in]{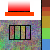}
} \subfloat[$\lambda=0.0500,
i=2$]{\includegraphics[width=1in]{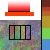}
} \subfloat[$\lambda=0.0500,
i=3$]{\includegraphics[width=1in]{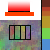}
} \subfloat[$\lambda=0.0500,
i=4$]{\includegraphics[width=1in]{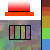}
} \subfloat[$\lambda=0.0500,
i=5$]{\includegraphics[width=1in]{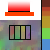}
}
\end{tabular}
\caption{Proposed edge aware nonlinear anisotropic diffusion
results on \emph{Challenge} image.  $\sigma_R=55$ for the
bilateral preconditioning in all cases.}
\label{vcd_challenge_impr}
\end{figure*}

The algorithms are also compared on the \emph{Challenge} image
(\figref{vcd_challenge} and \figref{vcd_challenge_impr}) while
varying the only two free parameters: $\lambda$, the diffusion
coefficient, and $i$, the number of filter iterations.

The edge aware version performs better than the original across
the gamut by preserving edges while still smoothing other areas.
As expected, reducing $\lambda$ reduces the tolerance for edges
and increases overall smoothing in both algorithms (nonlinearly).

However, it is interesting to note that, in the edge aware
version, extremely low values of $\lambda$ cause edge duplication.
This behavior is not important because such low values of
$\lambda$ do not make sense for discontinuity preserving smoothing
anyway.

\captionsetup[subfloat]{labelformat=empty,listofformat=empty,lofdepth}

\section{Edge Aware Modifications for\\ the Bilateral Filter} \label{bilateral_improve}

Consider the 4-connected image graph where pixels are nodes and
edges are weighted by the Euclidean distance between pixel colors.
Then there exists a shortest path on the graph between any two
pixels, and the length of this path is a representation of the
combined spatial and range distance between the pixels which takes
into account any discontinuities between them.

A traditional bilateral filter uses a spatial distance weight
multiplied by a color dissimilarity weight.  We propose to replace
this bilateral weighting of range and distance treated separately
by a single weighting of the shortest path distance on the image
graph.  This gracefully combines the two types of distances while
properly accounting for edge awareness.

If the shortest path distance between pixels $p$ and $s$ is
denoted by $d(p,s)$, then the modified bilateral filter can be
written as

\begin{equation}
J_s = \frac{1}{k} \sum_{p \in \Omega} g(d(p,s)) I_p
\label{sp_bilateral}
\end{equation}

Writing the filter in this way suggests some other interesting
filters, as well.  For example, consider using the maximum edge
distance instead of the sum of edge distances as the input to the
Gaussian for weighting.  Such a filter would eliminate gradients
in the source image, while preserving solid colored regions, which
might have interesting applications for image segmentation.

\subsection{Performance Analysis of Edge Aware\\ Bilateral Filter} \label{bilateral_perf}

The time complexity of processing an image with the regular
bilateral filter is $O(n d)$, where $n$ is the number of pixels
and $d$ is the number of pixels defined in the (square) local
neighborhood.  Note that $d = 4i^2$ in order to have a range of
influence of $i$, so it is significantly more expensive than
diffusion, which has complexity $O(n i)$ (for single processing
architectures).  In a massively parallel architecture, the
bilateral filter could be done in $O(d) = O(i^2)$ time and $O(1)$
communication, whereas diffusion would take $O(i)$ time and $O(i)$
communication.

Determining the path distance from a pixel $p$ to each neighbor $q
\in N_p$ is an instance of the \emph{single-source shortest-paths
}problem \cite{KLEINBERG06}.  It can be computed efficiently with
Dijkstra's algorithm \cite{DIJKSTRA59} in $O(m \log n)$ time
\cite{KLEINBERG06}, for a graph with $n$ nodes and $m$ edges.
Using the 4-connected image graph, each node contributes $\leq 2$
edges (eg, the edge above and to the left), so the total number of
edges will be $\leq 2d$ (note that $|N_p| \approx d$).  Thus, the
time complexity for solving the single-source shortest-paths
problem on a 4-connected image graph is $O(d \log d)$.  Computing
the bilateral weighted average would take $O(d)$ time already, so
this is only an additional factor of $O(\log d)$.

Because of the special circumstances of this problem, some
optimizations can be made such that the average case performance
is significantly better than the worst case performance.  First,
it should be noted that the spatial range (which determines the
size of $d$) is much less important, because there is no
mathematical reason to induce an arbitrary cutoff on the range of
influence.  Rather, it is more intuitive to specify the range in
terms of total path length.

If path length is used to specify the range of influence, then it
is only necessary to compute the distance to pixels $q \in N_p$ if
$d(p,q) < \tau_R$, where $\tau_R$ is a cutoff based on the range
standard deviation $\sigma_R$.  In these experiments, $\tau_R = 3
\sigma_R$.  Because paths are found in increasing order using
Dijkstra's algorithm \cite{DIJKSTRA59}, this means that
computation can terminate as soon as a path having length $ >
\tau_R$ is found.

In real images which do not usually contain large regions where
each pixel has \emph{exactly} the same intensity, there will only
be a constant number of pixels that can be reached from $p$
without exceeding the maximum path cost $\tau_R$ regardless of
spatial bandwidth.  Therefore, the average case time complexity is
$\lim_{d \to \infty} O(d \log d) = O(1)$.  However, to protect the
efficiency from suffering from a very high constant in homogeneous
regions, a spatial cutoff should still be used in addition to
$\tau_R$.

\subsection{Experimental Results of Edge Aware Bilateral Filter}

\begin{figure*}
\centering
\begin{tabular}{c}
\subfloat[Froud]{\includegraphics[width=2in]{froud.png}
}
\subfloat[Bilateral]{\includegraphics[width=2in]{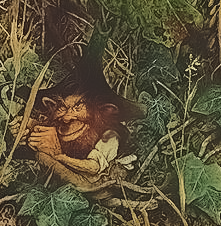}
}
\subfloat[Edge Aware Bilateral]{\includegraphics[width=2in]{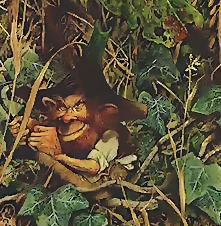} }\\
\subfloat[Paris]{\includegraphics[width=2in]{paris.png}
}
\subfloat[Bilateral]{\includegraphics[width=2in]{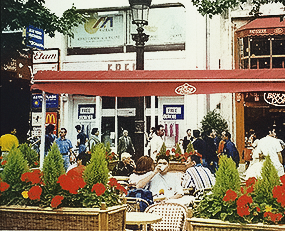}
}
\subfloat[Edge Aware Bilateral]{\includegraphics[width=2in]{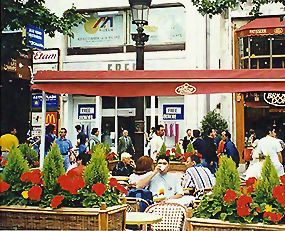} }\\
\subfloat[Nest]{\includegraphics[width=2in]{nest.png}
}
\subfloat[Bilateral]{\includegraphics[width=2in]{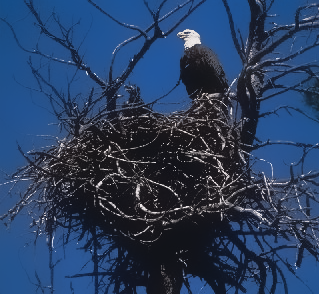}
}
\subfloat[Edge Aware Bilateral]{\includegraphics[width=2in]{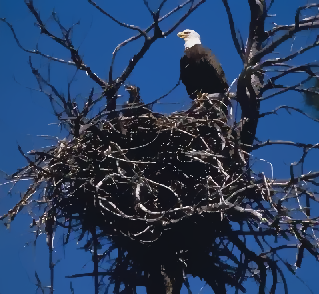} }\\
\subfloat[Egg]{\includegraphics[width=2in]{egg.png}
}
\subfloat[Bilateral]{\includegraphics[width=2in]{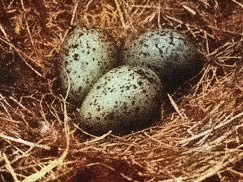}
}
\subfloat[Edge Aware Bilateral]{\includegraphics[width=2in]{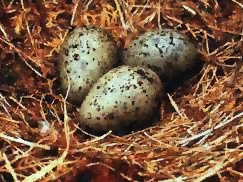} }\\
\end{tabular}
\caption{Comparison between bilateral filtered results on input
image (left column) using standard algorithm (middle column) and
edge aware modified algorithm (right column).  $\sigma_S=10$ and
$\sigma_R=55$ for all tests ($\sigma_R$ is used as the combined
spatial-range parameter in the edge aware version).}
\label{bilat_test_compare}
\end{figure*}

The original bilateral filter and the edge aware version are
compared on a few test images in \figref{bilat_test_compare}.  A
large spatial bandwidth is used specifically to highlight the
limitations of the bilateral filter.

All of the images filtered with the standard bilateral filter
exhibit a loss of saturation.  In the \emph{Froud} and \emph{Egg}
tests, this creates a somewhat surrealistic looking effect,
because foreground colors are blended with background colors.  In
the \emph{Paris} test, this is particularly noticeable in the
flowers which acquire an orange hue.  In the \emph{Eggs} image,
the color of the straw noticeably bleeds into the eggs.  The
bilateral filter does a fairly good job on the \emph{Nest} image
(due to the extremely simple histogram), but there is still some
visible cloudy gray effects in the sky to the right and left of
the nest (as well as desaturation of the sky).

\begin{figure*}
\centering
\begin{tabular}{c}
\subfloat[][$\sigma_R=20$,$i=1$]{\includegraphics[width=1in]{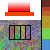}
}
\subfloat[][$\sigma_R=35$,$i=1$]{\includegraphics[width=1in]{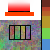}
}
\subfloat[$\sigma_R=50$,$i=1$]{\includegraphics[width=1in]{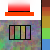}
}
\subfloat[$\sigma_R=65$,$i=1$]{\includegraphics[width=1in]{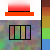}
}
\subfloat[$\sigma_R=80$,$i=1$]{\includegraphics[width=1in]{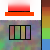} }\\
\subfloat[$\sigma_R=20$,$i=2$]{\includegraphics[width=1in]{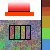}
}
\subfloat[$\sigma_R=35$,$i=2$]{\includegraphics[width=1in]{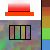}
}
\subfloat[$\sigma_R=50$,$i=2$]{\includegraphics[width=1in]{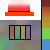}
}
\subfloat[$\sigma_R=65$,$i=2$]{\includegraphics[width=1in]{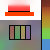}
}
\subfloat[$\sigma_R=80$,$i=2$]{\includegraphics[width=1in]{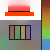} }\\
\subfloat[$\sigma_R=20$,$i=3$]{\includegraphics[width=1in]{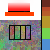}
}
\subfloat[$\sigma_R=35$,$i=3$]{\includegraphics[width=1in]{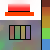}
}
\subfloat[$\sigma_R=50$,$i=3$]{\includegraphics[width=1in]{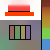}
}
\subfloat[$\sigma_R=65$,$i=3$]{\includegraphics[width=1in]{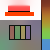}
}
\subfloat[$\sigma_R=80$,$i=3$]{\includegraphics[width=1in]{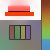} }\\
\end{tabular}
\caption{Proposed edge aware bilateral filtered results on
\emph{Challenge}, using $i$ filter iterations and combined
spatial-range parameter $\sigma_R$.} \label{bpds_test1}
\end{figure*}

In contrast, using the edge aware improved version based on
shortest path distance, colors maintain their bright saturation in
all of the tests in \figref{bilat_test_compare}.  Note that, in
general, even better results could probably have been obtained by
using a smaller bandwidth and more iterations (as is done in
\figref{bpds_test1}).

\begin{figure*}
\centering
\begin{tabular}{c}
\subfloat[$\sigma_R=20$,$\sigma=5$]{\includegraphics[width=1in]{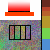}}
\subfloat[$\sigma_R=35$,$\sigma=5$]{\includegraphics[width=1in]{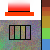}}
\subfloat[$\sigma_R=50$,$\sigma=5$]{\includegraphics[width=1in]{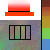}}
\subfloat[$\sigma_R=65$,$\sigma=5$]{\includegraphics[width=1in]{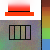}}
\subfloat[$\sigma_R=80$,$\sigma=5$]{\includegraphics[width=1in]{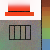}}\\
\subfloat[$\sigma_R=20$,$\sigma=10$]{\includegraphics[width=1in]{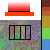}}
\subfloat[$\sigma_R=35$,$\sigma=10$]{\includegraphics[width=1in]{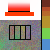}}
\subfloat[$\sigma_R=50$,$\sigma=10$]{\includegraphics[width=1in]{bilat_test1_sigmaR_50_sigmaS_10.png}}
\subfloat[$\sigma_R=65$,$\sigma=10$]{\includegraphics[width=1in]{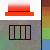}}
\subfloat[$\sigma_R=80$,$\sigma=10$]{\includegraphics[width=1in]{bilat_test1_sigmaR_80_sigmaS_10.png}}\\
\subfloat[$\sigma_R=20$,$\sigma=15$]{\includegraphics[width=1in]{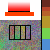}}
\subfloat[$\sigma_R=35$,$\sigma=15$]{\includegraphics[width=1in]{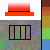}}
\subfloat[$\sigma_R=50$,$\sigma=15$]{\includegraphics[width=1in]{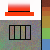}}
\subfloat[$\sigma_R=65$,$\sigma=15$]{\includegraphics[width=1in]{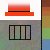}}
\subfloat[$\sigma_R=80$,$\sigma=15$]{\includegraphics[width=1in]{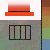}}\\
\end{tabular}
\caption{Standard bilateral filtered results on \emph{Challenge}
image.} \label{bilat_test1}
\end{figure*}

The algorithms are also compared on the \emph{Challenge} image.
In \figref{bilat_test1}, the original bilateral filter is tested
across a range of free parameter values.  The free parameters are
the usual $\sigma_R$ (standard deviation in the range dimension)
and $\sigma$ (standard deviation in the spatial dimension).  The
effects of color cancellation are very prominent across all
parameter ranges, and the noisy colored region is never
satisfactory smoothed out.

The results using a range of parameters with the edge aware
version is shown in \figref{bpds_test1}.  Note that these
parameters are completely different from the free parameters of
the original bilateral filter -- $\sigma_R$ is the combined
spatial-range parameter which is used on shortest path length.
The spatial parameter is essentially irrelevant -- we simply use a
large value so that all path-lengths go to zero before reaching
the boundary.  However, we do introduce a new parameter $i$,
number of iterations, which can be used to tweak the `spatial'
influence somewhat.

\section{Edge Aware Modifications \\for Mean Shift} \label{ms_ea}

Because the mean shift filter also uses a bilateral weighted
average, the computation of the mean shift vector can be done in
an edge aware way by simply adopting the same modification that
was proposed for the bilateral filter \eqnref{ms_bilateral}.
However, we found that this additional computation does not
improve the filtering result because the spatial bandwidth is
typically small enough that multiple similarly colored
distributions are not likely to occur within the support.

The more critical violation of edge awareness that that we wanted
to fix occurs when a concave feature causes the mean to shift off
of the feature.  In order to fix this problem, we propose to
compute the spatial component of the mean shift as normal, but to
shift to a location that is a compromise between the mean shift
vector and the image feature.  There are many conceivable ways to
accomplish this, and the method we use is by no means optimal, but
it works well in all of our tests.

Specifically, we begin by defining the set $S$ to contain all
pixels in the image with similar color to the current mean,

\begin{equation}
S = \left \{ p \in \Omega : \frac{||I_p - s_i^R ||}{h_r^2} < \tau
\right \}
\end{equation}

Then, if the next arithmetic spatial mean is $\mu$, we shift to
the spatial point $\hat p$ defined by

\begin{equation}
\hat p = \argmin_{p \in S} (p-\mu)^2
\end{equation}

We found that using $\tau=0.5$ works well in general, and we use
this value for all of our experiments.

\subsection{Performance Analysis of Edge Aware Mean Shift}

Finding the mode for each pixel requires $O(d i)$ time, where $d$
is the number of pixels in the local neighborhood (determined by
$h_s$) and $i$ is the number of iterations.  Because the mean
shift procedure is guaranteed to converge \cite{COMANICIU02}, $i$
is finite (for real images, the average number of iterations is
almost always somewhere in the range of $[1.1,1.8]$ using the
optimization mentioned below).

One of the optimizations proposed for the mean shift filter by
Christoudias et. al \cite{CHRISTOUDIAS02} is to
store the found mode for each pixel that falls on a mean shift
path.  Then, if another mean shift path falls upon a pixel that
was previously on another path, the new mean shift path uses the
previously found mode.

This makes the algorithm order-dependent and no longer guarantees that each pixel will be assigned to the
correct mode of the kernel density estimate, but they found it to produce very good results about 5-6 times faster (although algorithmic time complexity is unchanged).  This optimization is used in our implementation of mean shift as well.

\subsection{Experimental Results of Edge Aware Mean Shift}

\begin{figure*}
\centering
\begin{tabular}{c}
\subfloat[Froud]{\includegraphics[width=2in]{froud.png}
} \subfloat[Mean
Shift]{\includegraphics[width=2in]{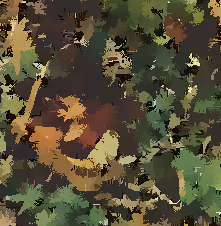}
}
\subfloat[Edge Aware Mean Shift]{\includegraphics[width=2in]{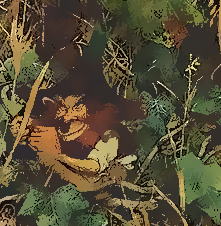} }\\
\subfloat[Paris]{\includegraphics[width=2in]{paris.png}
} \subfloat[Mean
Shift]{\includegraphics[width=2in]{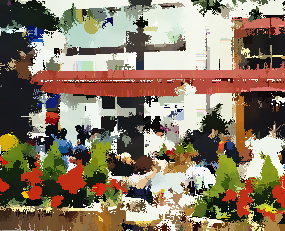}
}
\subfloat[Edge Aware Mean Shift]{\includegraphics[width=2in]{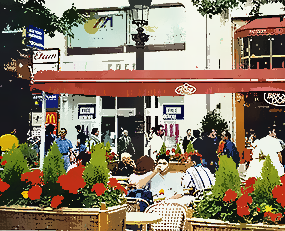} }\\
\subfloat[Nest]{\includegraphics[width=2in]{nest.png}
} \subfloat[Mean
Shift]{\includegraphics[width=2in]{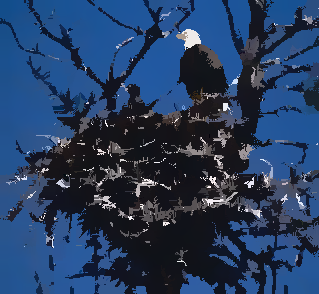}
}
\subfloat[Edge Aware Mean Shift]{\includegraphics[width=2in]{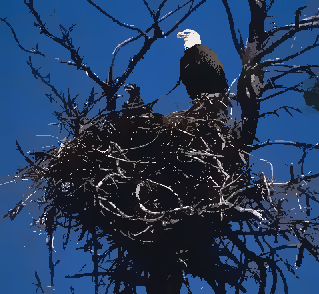} }\\
\subfloat[Egg]{\includegraphics[width=2in]{egg.png}
} \subfloat[Mean
Shift]{\includegraphics[width=2in]{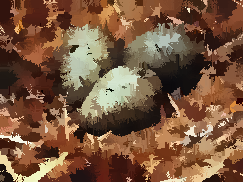}
}
\subfloat[Edge Aware Mean Shift]{\includegraphics[width=2in]{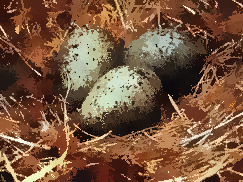} }\\
\end{tabular}
\caption{Comparison between mean shift filtered results on input
image (left column) using standard algorithm (middle column) and
edge aware modified algorithm (right column).  $h_s=11$ and
$h_r=55$ in all tests.} \label{ms_test_compare}
\end{figure*}

\begin{figure*}
\centering
\begin{tabular}{c}
\subfloat[$h_s=3$,$h_r=30$]{\includegraphics[width=1in]{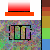}
}
\subfloat[$h_s=5$,$h_r=30$]{\includegraphics[width=1in]{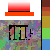}
}
\subfloat[$h_s=7$,$h_r=30$]{\includegraphics[width=1in]{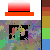}
}
\subfloat[$h_s=9$,$h_r=30$]{\includegraphics[width=1in]{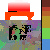} }\\
\subfloat[$h_s=3$,$h_r=60$]{\includegraphics[width=1in]{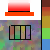}
}
\subfloat[$h_s=5$,$h_r=60$]{\includegraphics[width=1in]{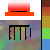}
}
\subfloat[$h_s=7$,$h_r=60$]{\includegraphics[width=1in]{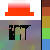}
}
\subfloat[$h_s=9$,$h_r=60$]{\includegraphics[width=1in]{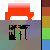} }\\
\subfloat[$h_s=3$,$h_r=90$]{\includegraphics[width=1in]{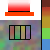}
}
\subfloat[$h_s=5$,$h_r=90$]{\includegraphics[width=1in]{ms_test1_hr_90_hs_5.png}
}
\subfloat[$h_s=7$,$h_r=90$]{\includegraphics[width=1in]{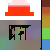}
}
\subfloat[$h_s=9$,$h_r=90$]{\includegraphics[width=1in]{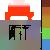}
}
\end{tabular}
\caption{Standard mean shift filtered result on \emph{Challenge}
test image.} \label{ms_test1}
\end{figure*}

\begin{figure*}
\centering
\begin{tabular}{c}
\subfloat[$h_s=3$,$h_r=30$]{\includegraphics[width=1in]{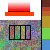}
}
\subfloat[$h_s=5$,$h_r=30$]{\includegraphics[width=1in]{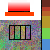}
}
\subfloat[$h_s=7$,$h_r=30$]{\includegraphics[width=1in]{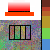}
}
\subfloat[$h_s=9$,$h_r=30$]{\includegraphics[width=1in]{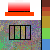} }\\
\subfloat[$h_s=3$,$h_r=60$]{\includegraphics[width=1in]{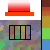}}\hspace{0.3pt}
\subfloat[$h_s=5$,$h_r=60$]{\includegraphics[width=1in]{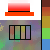}}
\subfloat[$h_s=7$,$h_r=60$]{\includegraphics[width=1in]{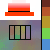}}
\subfloat[$h_s=9$,$h_r=60$]{\includegraphics[width=1in]{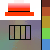}}\\
\subfloat[$h_s=3$,$h_r=90$]{\includegraphics[width=1in]{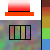}
}
\subfloat[$h_s=5$,$h_r=90$]{\includegraphics[width=1in]{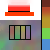}
}
\subfloat[$h_s=7$,$h_r=90$]{\includegraphics[width=1in]{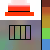}
}
\subfloat[$h_s=9$,$h_r=90$]{\includegraphics[width=1in]{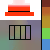}
}
\end{tabular}
\caption{Proposed edge aware mean shift filtered result on
\emph{Challenge} test image.} \label{ms_impr}
\end{figure*}

A comparison of the filtered results is shown in \figref{ms_test_compare}.  Figs.  \ref{ms_test1} and \ref{ms_impr} show results on the Challenge image using the original and modified algorithms, respectively.

\FloatBarrier

The most notable difference between the original and edge aware
modified filter results shown in \figref{ms_test_compare} is that
the edge aware version significantly cuts down on the ragged
edges.  A better looking filtered image could have been produced
by using a smaller spatial bandwidth, but that changes the scale
that is used to find modes in the kernel density estimate.  The
purpose of this test was to allow larger spatial bandwidths to be
used to find large scale modes in homogeneous regions while still
preserving fine detail in other regions.  The modified filter
accomplishes this.

In the \emph{Challenge } test, the original algorithm creates many
unsightly black artifacts (\figref{ms_test1}), and also causes
some of the white pixels to get assigned into the red block at
larger spatial bandwidths.  The edge aware version preserves the
strong black boundary lines while still clustering the underlying
colors according to the kernel density estimate
(\figref{ms_impr}).

As the scale grows larger, the edge aware version becomes
progressively less ``edge aware'', and eventually performs
blurring across boundaries.  A method for preventing this effect
was mentioned in \secref{ms_ea}, but it was specifically not used
because it disturbs the kernel density estimate too much.

\section{Conclusions}

Several successful approaches to the discontinuity preserving
smoothing problem have been reviewed, illustrating an interesting
conceptual dichotomy between viewing it as a global optimization
problem vs. a class of algorithms related to diffusion.  The
global optimization approaches all minimize the same fundamental
form using very different methods.  Variable conductance diffusion
is a direct application of physical diffusion, and anisotropic
diffusion is a simple variant of variable conductance diffusion.
The bilateral filter is a non-iterative approximation to
diffusion, and we showed that the mean shift filter is an
iterative generalization of the bilateral filter in higher
dimensions.

We showed that edge awareness is a desirable property for unbiased
discontinuity preserving smoothing for noise reduction, and
demonstrated examples where each of the above algorithms violated
the principle of edge awareness resulting in degraded image
quality: the diffusion algorithm originally had problems
maintaining edges after successive iterations, the bilateral
filter had issues with color bleeding and loss of saturation
(color cancellation), and the mean shift filter created
significant artifacts for non-convex colored regions when using
large spatial bandwidths.

Finally, we proposed individual modifications to each algorithm to
account for edge awareness and showed how these modifications
could be implemented with little or no change in algorithm
complexity.  After making the changes, visual improvements were
easily visible in the smoothed results.


\ifCLASSOPTIONcaptionsoff
  \newpage
\fi

\newpage

\end{document}